# NEURAL NETWORK TO IDENTIFY INDIVIDUALS AT HEALTH RISK

Tanja Magoc[1] and Dejan Magoc[2]

[1]Center for Bioinformatics and Computational Biology, University of Maryland, USA
`tmagoc@umiacs.umd.edu`
[2]Department of Health Studies, Eastern Illinois University, USA
`dmagoc@eiu.edu`

## ABSTRACT

*The risk of diseases such as heart attack and high blood pressure could be reduced by adequate physical activity. However, even though majority of general population claims to perform some physical exercise, only a minority exercises enough to keep a healthy living style. Thus, physical inactivity has become one of the major concerns of public health in the past decade. Research shows that the highest decrease in physical activity is noticed from high school to college. Thus, it is of great importance to quickly identify college students at health risk due to physical inactivity. Research also shows that the level of physical activity of an individual is highly correlated to demographic features such as race and gender, as well as self motivation and support from family and friends. This information could be collected from each student via a 20 minute questionnaire, but the time needed to distribute and analyze each questionnaire is infeasible on a collegiate campus. Thus, we propose an automatic identifier of students at risk, so that these students could easier be targeted by collegiate campuses and physical activity promotion departments. We present in this paper preliminary results of a supervised backpropagation multilayer neural network for classifying students into at-risk or not at-risk group.*

## 1. INTRODUCTION

Physical inactivity is one of the main causes of diseases such as heart attack and high blood pressure, and therefore it has become one of the biggest concerns of public health in the past decade [1],[2]. Even though many people claim that they exercise, only a very small percent of population is physically active at the level high enough to reduce the risk of diseases [3]. According to the American Heart Association and the American College of Sports Medicine, to keep a healthy living style, a person should exercise at least five times a week for 30 minutes at a moderate rate (e.g., walk), or at least three times a week for 20 minutes at a vigorous rate (e.g., run), or a combination of vigorous and moderate activities that result in an equivalent effort [4]. However, studies show that even though 75% of American population states to be physically active, only 25% of Americans satisfy the minimum recommendations [5].

Several studies have shown that there is a high correlation between the level of physical activity and demographic characteristics such as race and gender. Moreover, it has been shown that self-motivation, self perception of one's current physical and psychological health, support from friends and family, and previous physical engagement have a great impact on a person's activity level [6],[7],[8],[9],[10]. This information could be collected via a 20 minutes questionnaire from any individual, and then used to predict which individuals are likely to be under health risks due to physical inactivity. Once identified, the individuals at risk could be easier targeted by physical activity promotion departments and in future research involving physical activity and health. However, time to distribute the questionnaire to a large population and to analyze the results makes this process infeasible in reality. Currently, studies that attempt to find a general formulae for increasing physical activity of inactive or not sufficiently active individuals consist of attracting a small number of students to participate in surveys and providing tips to all participants how to improve health and physical activity. However, only a small number of inactive students is reached via these surveys since only limited amount of human power is available to distribute and analyze surveys, and there is no developed method to target individuals at risk, so all students (physically inactive as well as physically active) are targeted equally.

Since only a small number of inactive individuals is usually reached by physical activity studies, it is important to first quickly identify individuals at risk and then target this group of individuals in further studies and promotion programs. In order to identify individuals at risk, it is necessary to examine each individual, which is impossible by current method of distributing and analyzing questionnaires. Thus, this process needs to be automated and widely available.

Some aspects of health monitoring and providing tips for physical activities have been automated and available online. These tools include calculating body mass index [11], calculating real body age (regardless of biological age) based on the history of diseases and life style of an individual [12], and providing simple physical activity logs [13]. Moreover, a study was delivered to determine the impact of providing physical activity advice to elderly people via an automated computer system [14]. However, these tools only determine the current state of one's body and do not detect the reasons for possibly unhealthy life style, and they do not predict future health risk of an individual.

For this reason, we propose to develop a computerized questionnaire (for easy distribution) and the first automated identification of individuals at risk based on demographic and self-reported characteristics of in individual, which are provided via the questionnaire.

We developed a supervised multilayer backpropagation neural network to identify individuals at health risk based on demographic characteristics and self perception about physical activity. The neural network was trained on a set of data collected from 146 students at a predominantly Hispanic collegiate institution, and its accuracy was tested using the 5-fold cross-over validation technique. Using the available data, the developed neural network was able to classify correctly 79.5% of individuals. Even though the current neural network will enable targeting a large number of physically inactive students, when data from more individuals are available, the neural network will be re-trained and enhanced, and is expected to produce even better results.

## 2. COLLEGIATE SETTING FOR PHYSICAL ACTIVITY

Research shows that the most drastic decrease in physical activity occurs from high school to college and beyond [15]. Thus, it is important for college students to develop good physical activity habits since these habits transfer to later years of their lives [16],[17]. Campus activity boards often promote physical activity through inexpensive classes where a high support from peers and instructors is available, free gym membership, different sports competitions, as well as other promotions (such as free personal trainer or gift certificate) for participation in physical activity studies. However, most of these promotions usually attract students that are already physically active; thus ineffectively selecting research subjects and wasting resources on students who do not really need these resources (e.g., personal trainer or motivation from peers). Thus, identifying students that are likely to be at risk of being not physically active enough is crucial.

## 3. COLLECTED DATA

Since collegiate setting is the right place to start targeting individuals at health risk due to physical inactivity, we collected data from 146 students at a collegiate institution. The students were recruited from different classes representing diverse ethnic groups, gender, major area of study, and year in college. Each participant completed a 20-minute survey providing all information used in this study.

Each individual self-reported his/her *demographic factors* including gender and race. More females (62%) participated in the survey than males (38%). Since the survey was done at a predominantly Hispanic institution, it is not surprising that majority of participants were of Hispanic origin (82%). Besides Hispanics, several other ethnic groups were present in the study including Caucasians, African Americans, Native Hawaiians, American Indians, and Asians. However, with only 18% of participants belonging to races other than Hispanic, there were not enough samples to consider each of non-Hispanic ethnicities individually; thus, all participants were classified either as Hispanic or non-Hispanic. The summary of demographic characteristics is presented in the Table 1.

Table 1. Demographic characteristics of participants.

|  | **Male** | **Female** |
|---|---|---|
| Hispanic | 46 | 74 |
| Non-Hispanic | 10 | 16 |

Each participant also self-reported the *major area of study* at the college. It is expected that students majoring in health and sports related fields, such as kinesiology, pre-nursing, and physical therapy, are more aware of the impact of physical activity on health and are therefore more active than their peers majoring in areas not related to health and sport studies, such as education, science, and engineering. The hypothesis is supported by our sample of students with about two-thirds of sports and health related majors not being at risk of inactivity, and more than one-half of not sports and health related majors being under the risk of inactivity. Each participant was therefore classified by his/her major as either sports related student or not sports related student. The summary of participants' answers is provided in the Table 2.

Table 2. Major area of study and the risk of physical inactivity.

|  | **Sports related** | **Not sports related** |
|---|---|---|
| At risk | 33 | 24 |
| Not at risk | 70 | 19 |

Moreover, the participants self-reported *perceptions of their current physical and psychological health* by selecting one of five available options: excellent, good, fair, poor, and very poor. Most of the participants self-reported their physical health to be between good and fair, and psychological health to be good. Furthermore, the participants self-reported the *perception of their diet quality* with majority reporting their diet quality to be fair or good. The complete summary of the participants' answers is given in the Table 3.

Table 3. Physical and psychological health and diet quality of participants as recorded by participants' subjective perception.

|  | **Physical health** | **Psychological health** | **Diet quality** |
|---|---|---|---|
| Excellent | 21 | 32 | 8 |
| Good | 59 | 77 | 43 |
| Fair | 50 | 30 | 67 |
| Poor | 16 | 7 | 23 |
| Very poor | 0 | 0 | 5 |

The participants completed the *Self-Efficacy for Exercise Behavior Scale Assessment* [18], which measures the individual's readiness to overcome obstacles (e.g., tiredness, large amount of work, not accomplishing set physical goals) in order to exercise. Moreover, the participants reported the *importance of setting aside time for exercise* in their schedule and following set goals. Participants also reported the *expectations from the exercise* and the importance of each of the expectations. For each of the offered expectations (e.g., be healthy, lose weight, have fun), a participant could choose whether it is a reason for his/her exercise and how important it is for him/her. Furthermore, the participants completed the *Exercise Habits Scale Assessment* [19], which measures the support and motivation participants receive form family and friends to exercise. The motivation and support include features such as exercising together with another individual, receiving reminders from an individual to exercise, or discussing exercises. For each of the questions, the participants were able to provide the extent to which an activity (e.g., exercise together with a friend) is satisfied by selecting one of five options that best describes the provided answer: really high, high, medium, low, or really low. Each assessment (e.g., self-efficacy for exercise behavior assessment) was then represented by one value, which was obtained by averaging answers to all questions belonging to that assessment. The summary of the participants' answers is provided in the Table 4.

Table 4. Self-efficacy to overcome obstacles in order to exercise, importance of physical activity, expectations from physical activity, and support and motivation received from family and friends as reported by participants.

| | Self-efficacy | Importance | Expectations | Support |
|---|---|---|---|---|
| Really high | 49 | 5 | 40 | 25 |
| High | 61 | 35 | 48 | 44 |
| Medium | 29 | 58 | 36 | 50 |
| Low | 6 | 47 | 17 | 22 |
| Really low | 1 | 1 | 5 | 5 |

Finally, the participants filled out a short version of the *International Physical Activity Questionnaire* [20], which is a self-reporting measurement of the level of physical activity the individual performed during the last seven days. The physical activities are divided into two main categories based on the intensity of the activity: vigorous and moderate. A moderate activity is considered to be a planned physical activity done continuously for 30 minutes or in intervals of at least 10 minutes. It mildly elevates the heart rate and breathing, and includes activities, such as low-impact exercise/strength classes, walking, cycling less than 3 miles, recreational sports, and hiking among others. A vigorous activity is a physical exercise done continuously for at least 20 minutes. It elevates heart rate and breathing to the level when the individual is not able to hold a conversation while exercising, and includes activities such as running or jogging, high-intensity aerobic classes, competitive sports, swimming laps, and cycling for more than 3 miles. The participants reported the number of days and minutes per day that they spent in the last seven days on vigorous and on moderate physical activities [21].

## 4. NEURAL NETWORK APPROACH TO IDENTIFYING INDIVIDUALS AT HEALTH RISK

Currently available data obtained in numerous studies about physical activities suggest high correlation between certain demographic and self-perception factors and the level of an individual's exercise. However, dependencies among different factors and the level to which each factor impacts an individual's readiness and commitment to exercise are not precisely known. Due to these uncertainties, we employed a machine learning approach (see e.g., [22],[23]) to capture important relationships among features present in the existing data. The inferred relationships are then used to identify individuals at health risk using only a few demographic and self-reported characteristics of individuals.

Using the information collected from 146 students (as described in the previous section), we trained a backpropagation neural network (NN) with eight input variables, one hidden layer with 19 nodes, and the output layer with two nodes. The input variables include the following:

- *Gender*: this variable could take two values: male and female.
- *Hispanic*: this variable could take two values: yes and no, describing whether the individual is Hispanic or not, respectively. This distinction was made since studies have shown that Hispanic population exhibits different attitudes towards physical activities from those exhibited by, for example, Caucasian people [7]. Since not enough data were available to train the neural network on different ethnicities among non-Hispanics, further distinction among ethnicities was not included, but will be included in further improvements of the proposed NN method.
- *Major*: this variable could take two values: sport related and not sport related. This distinction was made because students majoring in sport or health related studies are more likely to be aware of physical activity importance and therefore, on average, exercise more than their peers who major in other disciplines.
- *Physical health*: this variable is a self-reported perception of the individual, and could take any of the five values: excellent, good, fair, poor, and very poor.
- *Psychological health*: this variable is a self-reported perception of the individual, and could take any of the five values: excellent, good, fair, poor, and very poor.
- *Diet*: this variable is a self-reported perception of the individual, and could take any of the five values: excellent, good, fair, poor, and very poor.
- *Self-efficacy*: this variable is a summary of an individual's answers on the self-efficacy questionnaire. Since each question allowed a participant to express the level of self-efficacy in the range 1-5 (1 meaning 'low' and 5 meaning 'high'), the values were averaged, and the average in the interval (4.00,5.00] was reported as "really high", the average in the interval (3.00,4.00] as "high", etc. The self-efficacy variable could take one of five values: really high, high, medium, low, and really low.
- *Importance of exercise*: this variable represents how important it is for the individual to make time for exercise in his/her schedule and to accomplish the scheduled physical activity goals. The variable could take one of five values: really high, high, medium, low, and really low.
- *Expectations*: this variable is a summary of the individual's answers to the proposed expectations from exercise, and the frequency at which the person exercises to accomplish that particular expectation (e.g., to lose weight). For each provided expectation, an individual could select the importance of that expectation in the range 1-5, and the value of exercise to accomplish that goal in the range 1-3. Corresponding values were multiplied, and the products averaged. The obtained value was divided by 3 to map the possible values to the interval [1,5], so that the *expectations* variable could be defined by the same labels and in the same manner as the *self-efficacy* variable.
- *Support*: this variable is a summary of the individual's answers to the exercise habits scale assessment questionnaire. It was created similarly to the *self-efficacy* variable and could take one of the five values really high, high, medium, low, and really low.

The output layer of the neural network consists of two nodes: *risk* and *no risk*. Only one of these two nodes is *on* as a result of applying the neural network to data collected from a new individual. Depending on which node is *on*, the person is classified to be or not to be at health risk based on the self-reported characteristics.

A neural network was trained using free software package *weka* [24]. Since all collected data were represented as non-numeric data to *weka*, each variable that could take more than two values was represented by multiple input nodes, one for each value the variable could take. For example, the variable *importance of exercise* could take one of five values (really high, high, medium, low, and really low), thus five input nodes were designed for this particular variable. However, even though some variables could take one of five values, not all five values have shown up in the collected data sample, and thus, less nodes were used to represent such a variable. For example, no one reported his/her physical health to be *very poor*, thus the *physical health* variable used only four nodes in the developed neural network. Variables with multiple nodes, such as physical health and importance of exercise, would have only one of their nodes set *on* in each training or testing sample.

Furthermore, if a variable could take exactly two values, this variable was represented by only one node. This node was either *on* or *off*, representing two different values that the variable could take. With this representation, the input layer of the neural network developed for the collected data sample consisted of 25 nodes.

The hidden layer was automatically generated by *weka*. By default, *weka* generates

$$\frac{number\ of\ attributes\ +\ number\ of\ classes}{2}$$ nodes in the hidden layer, which

resulted in (36 input nodes + 2 output nodes)/2=19 nodes for our data sample. All nodes were designed as sigmoid nodes due to non-numeric nature of data, which is the *weka*'s default construction.

A fully connected neural network (i.e., a NN where each node from a previous layer is connected to each node of the next layer, see Figure 1) was trained using a backpropagation algorithm. Backpropagation allows a neural network to infer the error rate at each node in the hidden layer based on the error in classification, which is manifested in the output nodes. This approach is well-suited to adjust weights on edges coming out of hidden layer nodes as well as edges coming into the hidden layer nodes. Thus, the backpropagation algorithm enables reduction of the error at each node in the hidden layer, and therefore a better model with reduced final classification errors.

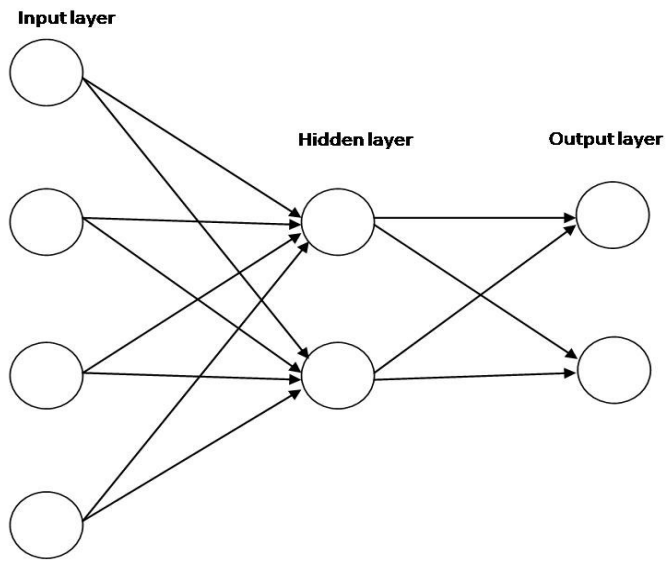

Figure 1. An example of a fully-connected multi-layer neural network with one hidden layer of nodes. Note that this figure is much smaller than the NN used in our project.

The learning rate in the neural network was initially set at 0.2 and decreased throughout training cycles. Total of 500 cycles were performed to train this neural network, which took 3.35 seconds for the given data set.

Since supervised learning was applied to train the neural network, the expected class for each individual was manually predicted. A person was classified to be not at risk if, in the last seven days, he/she reached at least one of two requirements suggested by the American Heart Association and the American College of Sports Medicine (i.e., exercise at least three times per week for at least 20 minutes at a vigorous rate, or at least five days per week for at least 30 minutes at a moderate rate). Moreover, some individuals did not reach any of these two recommendations directly, but performed an equivalent amount of exercise when combining the amount of vigorous and moderate activities during the week (e.g., two days of 30 minutes of vigorous activity and two days of 50 minutes of moderate activity). To decide which combinations of moderate and vigorous activities are equivalent or higher than the minimum exercise requirements, we calculated the metabolic equivalent of task (MET), which is the amount of energy used for each physical activity. The weekly MET of an individual was calculated as

$$MET = 4 \cdot m_d \cdot m_m + 8 \cdot v_d \cdot v_m,$$

where $m_d$ and $m_m$ are the number of days in the last seven days and average minutes per day that moderate activities were performed, respectively, and $v_d$ and $v_m$ are similarly days per week and average minutes per day, respectively, spent on vigorous activities. If the MET resulted above 600, the person was classified to be not at risk. The individuals that did not satisfy either of two recommended minimum levels of activity and did not spend energy equivalent to or higher than 600 METs were classified as being at risk.

## 5. RESULTS

We tested the developed neural network by using 5-fold cross-validation technique on the collected data. Combining results from all five runs, the neural network classified correctly 79.5% of test data.

Since the goal of the software is to identify individuals at risk, we consider true positive results to be the correct predictions of the students at risk and true negative results to be correct classification of students not at risk. Accordingly, the false negative results are students at risk incorrectly classified as not being at risk, and false positive are the students that are not at risk but were classified incorrectly as being at risk. The true positive (TP), true negative (TN), false positive (FP), and false negative (FN) rates are summarized in the Table 5.

Table 5. Accuracy of the prediction results.

| TP | TN | FP | FN |
|----|----|----|----|
| 0.83 | 0.74 | 0.26 | 0.17 |

There are several reasons for misclassification of the neural network. One of the reasons is that even though there is a high correlation between the level of physical activity and each of the features considered in this study, the correlation is not perfect, so there are some extreme cases (i.e., individuals) that do not follow common trends. Unfortunately, it is practically impossible to train a neural network to classify these individuals into the correct class within the current design of the neural network. However, designing another class for possible extreme cases might be an option to avoid misclassification of these individuals by marking them as individuals that should be evaluated further in order to be classified correctly. To train NN to recognize this additional class, we need to collect more data that contain information about individuals that clearly do not follow common trends.

Another reason for misclassification is that collected data is somewhat biased for certain variables, for instance, *race*. Since majority of participants were Hispanics, all the other races were considered as non-Hispanics, but there are cultural differences among non-Hispanic population, which might have impact on physical activity. However, the current sample does not contain enough training samples to train the neural network taking into consideration different non-Hispanic races. Moreover, there are none or only a few individuals in the sample that self-reported *very poor* diet, physical or psychological health. Thus, not enough training is possible in these "extreme" cases.

Finally, with the exception of demographic characteristics and students' major, the information used in this study is mostly a subjective perception of participants. Some variables (e.g., self-efficacy, support from family and friends) are constructed from multiple questions, and are therefore relatively objective values. However, other variables might be more subjective and not consistent throughout the sample. For example, an individual might weight 220lbs last year and 200lbs this year, so this individual might report his/her *physical health* to be good (compared to the last year); however, this person might still be overweight and thus objectively is not in the good physical health.

# 6. FUTURE WORK

The current neural network predictions will allow college campuses and physical activity promotion programs to target a large number of students at health risk due to physical inactivity. However, further improvements in the predictions are possible. We are currently collecting more data from students at different collegiate institutions. Having more data will allow better training of neural network. Moreover, we expect to collect data from students with wider variety of demographic characteristics, which would aid predictions in the non-Hispanic subgroups that are not adequately represented in our current data sample. Furthermore, larger number of samples would most likely contain more individuals reporting *very poor* health conditions, and aid neural network training for these extreme samples.

A more objective assessment of the physical and psychological states of each individual is preferable. For example, a more objective measure of physical health could be obtained by combining facts such as weight, height, and body mass index with the individual's subjective perception. A more objective measure of psychological health could be obtained by asking multiple questions describing possible stressful situations (e.g., whether the student have had enough sleep over nights) rather than just one question asking students to self-rate their psychological health.

It is also desirable to more objectively evaluate physical activities performed by individuals to determine whether a particular activity is of moderate or vigorous intensity. For example, while one person might consider running at 12mph to be a vigorous activity, it would not be considered a high intensity exercise by a person that is highly physically active. Since this information is used in supervised learning to determine the expected outputs (i.e., the class), it is important to have a consistent and correct interpretation of each physical activity. Furthermore, individuals do not tend to perform the same amount of exercise each week. Twenty-five individuals in this study reported their weekly physical activities for seven consecutive weeks. These reports show that 11 of these 25 participants (44%) would be considered *at-risk* one week and *not at-risk* in another week. However, seven of these 11 individuals showed a consistent behavior for six weeks (i.e., either being adequately active for six weeks and not adequately active in one week, or not being adequately active for six weeks and adequately active in one week). Thus, these individuals could be mislabeled if we consider only one week, but likely correctly classified if considered all seven weeks. Therefore, to more correctly classify these individuals in one of two groups, weekly activity logs should be collected over a few weeks and the average could be used for the final classification.

Since there are individuals that are on the borderline between two groups, a neural network with three classes will be developed: a class of individuals that are active above the minimum requirements, a class of individuals that are active but do not meet the minimum requirements (this group will include the individuals that are sufficiently active one week but not sufficiently active the next week), and the class of individuals that are not active at all.

Finally, once more data are collected and more diverse sample of population is reached, we will train a final neural network and develop a web-based tool to administer the questionnaire and immediately identify individuals at risk of being physically not active enough. We will make this program easily accessible to individuals that are not tech savy in order to improve well being of general population.

# 7. CONCLUSION

Majority of general population does not engage in enough physical activity to keep healthy life style. Physically inactive or very little active individuals are at high risk of health diseases such as heart attack and high blood pressure. A healthy living style is developed throughout life starting at childhood and especially through the entire collegiate years. Research shows that the highest decrease in the level of physical activity is denoted in the transition from high school to college. Thus, it is important to quickly identify college students at health risk in order to target these students in health research studies and physical activity programs.

Since intensive research shows that there is a high correlation between the level of physical activity and features such as race, gender, self-motivation, and support from family and friends, these features could be used to detect individuals at risk by collecting this information via a 20-minute questionnaire and analyzing the results. Currently, the surveys are written by individuals who are doing a particular study and are not uniform throughout the similar studies. Moreover, the results are often analyzed using descriptive statistical methods, and only summary results are reported at the end of the study without identifying and helping individuals at risk. Thus, even students that participate in surveys are often not aware at the end of the study that they might be at health risk or that they are not active enough. The main reason for not analyzing in detail each individual's survey answers is that it is time consuming and therefore infeasible to do manually.

To reduce time needed to collect and analyze the information, we propose to develop a web-based automated data collector and analyzer. In this paper, we presented a preliminary supervised backpropagation multilayer neural network to identify students at risk based on only a few demographic and self-reported characteristics, which classified correctly 79.5% of individuals. By collecting more data from students with diverse cultural backgrounds, and using these data to train the neural network, the correctness of the developed neural network will increase, and a freely available web application will be developed to allow quick and easy impact on health improvement in general population.

**Authors**

Tanja Magoc is a postdoctoral fellow in the Center for Bioinformatics and Computational Biology at the University of Maryland utilizing data mining techniques to analyze genomics and physical activities impact on developing diseases.

Dejan Magoc is an Assistant professor in the Department of Health Studies at the Eastern Illinois University working on relationships between behavioural change theories and physical activity as well as developing web based approaches to increase the level of physical activities in individuals.